\documentclass[10pt,twocolumn,letterpaper]{article}

\usepackage{iccv}
\usepackage{times}
\usepackage{epsfig}
\usepackage{graphicx}
\usepackage{amsmath}
\usepackage{amssymb}

\usepackage{subfigure}
\usepackage{mwe}
\usepackage{bbm}
\usepackage{booktabs}
\usepackage{algorithm}
\usepackage{algpseudocode}
\usepackage{color, colortbl}
\usepackage{amsthm}

\definecolor{Gray}{gray}{0.85}
\DeclareMathOperator*{\argmax}{arg\,max}

\theoremstyle{definition}
\newtheorem{definition}{Definition}[]

\usepackage[breaklinks=true,bookmarks=false,colorlinks=true]{hyperref}

\iccvfinalcopy 

\def\iccvPaperID{3184} 

\ificcvfinal\fi

\begin{document}

\title{\vspace{-3em}Walk in the Cloud: Learning Curves for Point Clouds Shape Analysis}

\author{Tiange Xiang\\
University of Sydney\\
{\tt\small txia7609@uni.sydney.edu.au}
\and
Chaoyi Zhang\\
University of Sydney\\
{\tt\small chaoyi.zhang@sydney.edu.au}

\and
Yang Song\\
University of New South Wales\\
{\tt\small yang.song1@unsw.edu.au}

\and
Jianhui Yu\\
University of Sydney\\
{\tt\small jianhui.yu@sydney.edu.au}

\and
Weidong Cai\\
University of Sydney\\
{\tt\small tom.cai@sydney.edu.au}
}

\maketitle
\ificcvfinal\thispagestyle{empty}\fi

\begin{abstract}
   Discrete point cloud objects lack sufficient shape descriptors of 3D geometries. In this paper, we present a novel method for aggregating hypothetical \textit{curves} in point clouds. Sequences of connected points (\textit{curves}) are initially grouped by taking guided walks in the point clouds, and then subsequently aggregated back to augment their point-wise features. We provide an effective implementation of the proposed aggregation strategy including a novel \textit{curve grouping} operator followed by a \textit{curve aggregation} operator. Our method was benchmarked on several point cloud analysis tasks where we achieved the state-of-the-art classification accuracy of 94.2\% on the ModelNet40 classification task, instance IoU of 86.8\% on the ShapeNetPart segmentation task and cosine error of 0.11 on the ModelNet40 normal estimation task. Our project page with source code is available at: \url{https://curvenet.github.io/}.
   
\end{abstract}

\section{Introduction}

The point cloud is a primary data structure in a string of indoor/outdoor computer vision applications. A large variety of 3D sensors (e.g. LiDAR sensors) are now able to capture real-world objects, and their projections to digital forms can be made by sampling discrete points on the surface. To reach a better understanding of the 3D targets, effective point cloud analysis techniques and methods are in great demand. With the thriving of deep learning, the pioneer works \cite{qi2017pointnet,qi2017pointnet++} and their followers \cite{liu2020closer,hu2020randla,yan2020pointasnl, bytyqi2020local,xu2020grid,hu2020jsenet,lin2020point2skeleton,SGGpoint} processed point clouds through well-designed neural networks to learn the latent mappings between input point coordinates and the ground truth labels. Differing from conventional 2D vision tasks, the points are usually in irregular and unordered forms, hence, effective design of feature aggregation and message passing schemes among point clouds still remains challenging.

Local feature aggregation is a basic operation that has been widely studied recently. For each key point, its neighborhood point features are first grouped by pre-defined rules (e.g. KNN). The relative position encodings between the query point and neighboring points will be computed subsequently and then passed into various point-based transformation and aggregation modules for local feature extraction. Although the above operations help depict local patterns to some extent, long-range point relations are neglected. While the non-local module \cite{wang2018non} provides a solution to aggregate global features, we argue that the global point-to-point mapping might still be insufficient to extract underlying patterns implied by the point cloud shapes. 


\begin{figure}
	\begin{center}
		\includegraphics[width=1\linewidth]{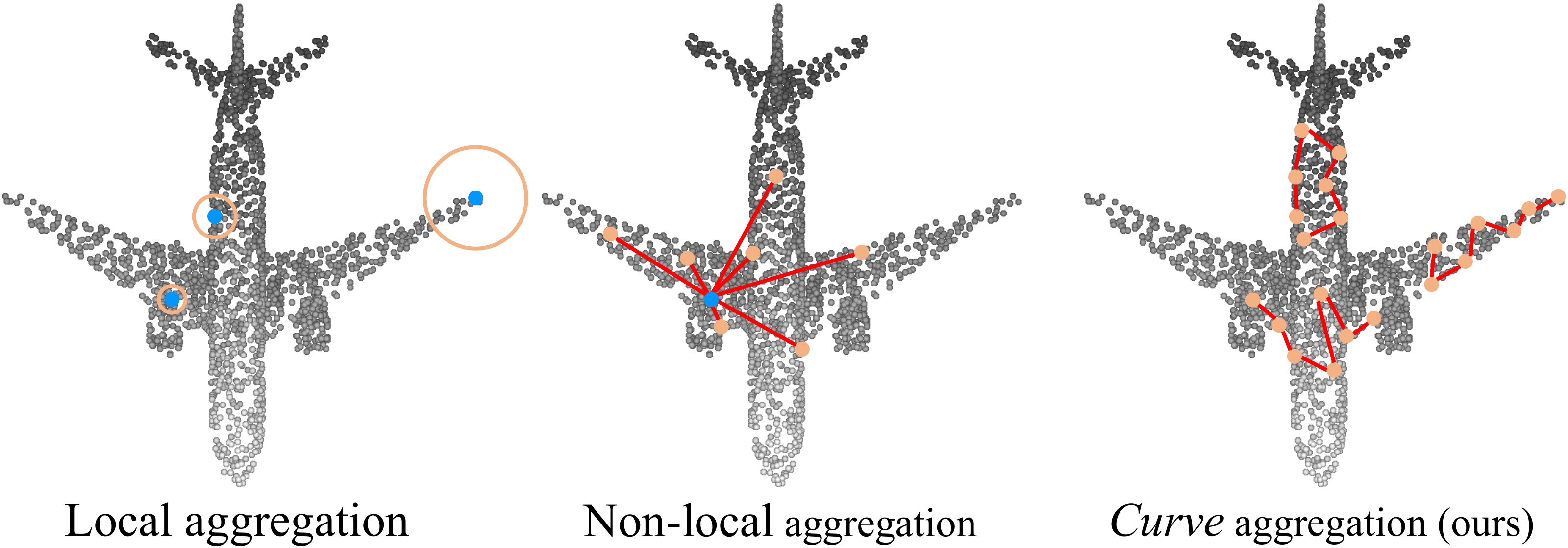}
	\end{center}
	\caption{\textbf{Common aggregations and the \textit{curve} aggregation}. Blue circles denote key points, orange circles denote query points or query range. In \textit{curve} aggregation, query features will be aggregated into all points.}
	\label{fig:1}
\end{figure}

To this end, we propose to improve the point cloud geometry learning through generating continuous sequences of point segments. We posit that such continuous descriptors are more adequate for depicting the geometry of point cloud objects, compared to popular existing local and non-local operators nowadays. We denote such continuous descriptors as \textit{curves}. By regarding a point cloud as an undirected graph, where the discrete points serve as the graph nodes and the neighbor point connections as the graph edges, a \textit{curve} can therefore be described as a walk in the graph. Figure \ref{fig:1} intuitively compares the local aggregation, non-local aggregation, and our \textit{curve aggregation} operators. In this paper, we first revisit the local feature aggregation in its general form and provide in-depth discussions on why long-range feature aggregation strategies are desired (Sec. \ref{rethink}). We then formulate our method formally by defining the curve grouping policy (Sec. \ref{cg}) and the aggregation between curve features and point features (Sec. \ref{ca}). A novel point cloud processing network, CurveNet, is constructed by integrating the proposed modules along with several basic building blocks into a ResNet \cite{he2016deep} style network. 




Our main contributions are three-fold: \textbf{(1)} We propose a novel feature aggregation paradigm for point cloud shape analysis. Sequences of points (\textit{curves}) are grouped and aggregated for better depiction of point cloud object geometries. A novel curve grouping operator along with a curve aggregation operator are proposed to achieve curve feature propagation. \textbf{(2)} We study potential drawbacks of grouping \textit{loops} and provide our solutions to alleviate them. Moreover, a dynamic encoding strategy is proposed so that \textit{curves} can contain richer information while suppressing potential \textit{crossovers}. \textbf{(3)} We embed the curve modules into a network named \textit{CurveNet}, which achieves state-of-the-art results on the object classification, normal estimation, and object part segmentation tasks. 


\section{Related Works}

\noindent
\textbf{3D point cloud processing.} One of the greatest challenges to point cloud analysis is processing unstructured representations. Starting from indirect representation transformation methods \cite{li2016fpnn, klokov2017escape, riegler2017octnet, lei2019octree} that first transform the point cloud into another representation (e.g. octree, kdtree) to ease the analysis difficulty, many recent works \cite{qi2017pointnet,zhang2020shape,hu2020randla} extract features from the raw point cloud directly.



As one of the pioneer direct approaches, PointNet/PointNet++ \cite{qi2017pointnet, qi2017pointnet++} utilize shared MLPs to learn point-wise features. Following them, recent works have extended the point-wise method to various directions, which include designing advanced convolution operations \cite{komarichev2019cnn, xu2020grid, wu2019pointconv, liu2019relation}, considering a wider neighborhood \cite{li2018so, zhao2019pointweb, wang2019dynamic, liu2019lpd}, and the adaptive aggregation of it \cite{hu2020randla, yan2020pointasnl, zhang2019pcan, yang2019modeling}. The success of the above methods is inseparable from the help of feature aggregation operators, which achieve the direct message passing of discrete points in deep networks.

Current feature aggregation operators can be generally classified into two categories: local feature aggregation and non-local feature aggregation. As a representative of local aggregation operator, \textit{EdgeConv} \cite{wang2019dynamic} learns semantic displacement between key points and their feature-space neighbors. The thrive of non-local aggregation operator starts from the non-local network \cite{wang2018non}, with which global features are transformed and aggregated together to learn many-to-one feature mappings. With the recent success of applying Transformer \cite{vaswani2017attention} in vision tasks, Guo \etal \cite{guo2020pct} designed a point cloud processing architecture comprised of simple Transformers. 





Beyond the local and non-local feature aggregation operators, we suggest that point cloud analysis can be better achieved with special consideration of shape segments, edges, and curves. By aggregating the additional \textit{curve} features, latent feature information can be enriched. 

\noindent
\textbf{Sampling techniques for 3D point cloud.} Sampling technologies aggregate indicative point patterns and are hence essential to all point cloud processing methods. Voxelization-based approaches \cite{qi2016volumetric, wu2016learning, wu20153d, maturana2015voxnet} transform the discrete point space into 3D grid (voxels), where the input point clouds behave as the discrete sampling on a continuous 3D space. However, the quality of such sampling is highly sensitive to the subdivision frequency. Similar to voxelization-based sampling methods, view-based approaches \cite{su2015multi, kanezaki2016rotationnet, xie2016deepshape} start from capturing 2D snapshots of point clouds from different angles, and make predictions based on the 2D images. The loss of spatial information is inevitable during such image samplings.

Advanced sampling methods from recent literatures overcome the above drawbacks, and obtain promising results on basic point cloud analysis tasks. GS-Net \cite{xu2020geometry} exploits an Eigen-Graph to group points with similar Euclidean distance and geometric information. PointASNL \cite{yan2020pointasnl} samples both adjacent and global points for a complete description of point cloud objects. Unlike the above methods, PAT \cite{yang2019modeling} models point clouds with the help of a Transformer \cite{vaswani2017attention} and learns the point sampling through a Gumbel-Softmax gate. RandLA-Net \cite{hu2020randla} reviews multiple sampling techniques and adopts the random sampling for efficiently processing large scale point clouds. In a concurrent work, MeshWalker \cite{lahav2020meshwalker} also experiments on taking random walks on mesh surfaces for better mesh analysis. Differing from all existing sampling methods, we take guided walks to group contiguous segments of points as curves, which contain rich information describing object shapes and geometry.

\section{Methods}
In this section, we present the proposed operators to group and aggregate curves for any point cloud $\mathbf{P}=\{\mathbf{p}\}$ and its point-wise features $\mathbf{F}=\{\mathbf{f}\}$. As aforementioned, a curve represents a connected sequence of points in the point cloud, and can be formally defined as:

\begin{definition}
\textit{\textbf{Curves in point cloud.}} Given $\mathbf{P}$, $\mathbf{F}$ and an isomorphic graph $\mathbf{G}=(\mathbf{F}, \mathbf{E})$ with the connectivity $\mathbf{E}$ computed by the KNN algorithm on $\mathbf{P}$. A curve $\mathbf{c}$ with length $l$ in feature space, is generated as a sequence of point features in $\mathbf{F}$, such that $\mathbf{c}=\{\mathbf{s}_1,\cdots,\mathbf{s}_l|\mathbf{s} \in \mathbf{F}\}$. To group curves, we consider a walk policy $\pi$ defined on the isomorphic graph $\mathbf{G}$ that starts a walk (curve) from a starting point $\mathbf{s}_1$ and transits for $l$ steps. 
\end{definition}


\subsection{Rethinking Local Feature Aggregation} \label{rethink}

The general purpose of local feature aggregation is to learn the underlying patterns within a local space of $k$ elements. For each point $\mathbf{p}$, the neighborhood $N=\{\mathbf{p}^1,\cdots,\mathbf{p}^k\}$ is first grouped by a deterministic rule, and KNN is the most frequently used grouping algorithm \cite{hu2020randla,yan2020pointasnl,liu2020closer,zhang2020shape} due to its computational efficiency. Then, pair-wise differences $\Phi$ between every two elements in $N$ are computed and stacked together. Finally, shared MLPs are employed to further aggregate the computed encodings leading to the locally aggregated features $\mathbf{g}$. Formally, the above local feature aggregation process can be formulated as:
\begin{equation} \label{la}
    \mathbf{g} = \texttt{pooling}(\{\texttt{MLP}(\Phi(\mathbf{f}, \mathbf{f}^j)| \mathbf{f}^j\in N)\}).
\end{equation}

\begin{figure}[t]
	\begin{center}
		\includegraphics[width=1\linewidth]{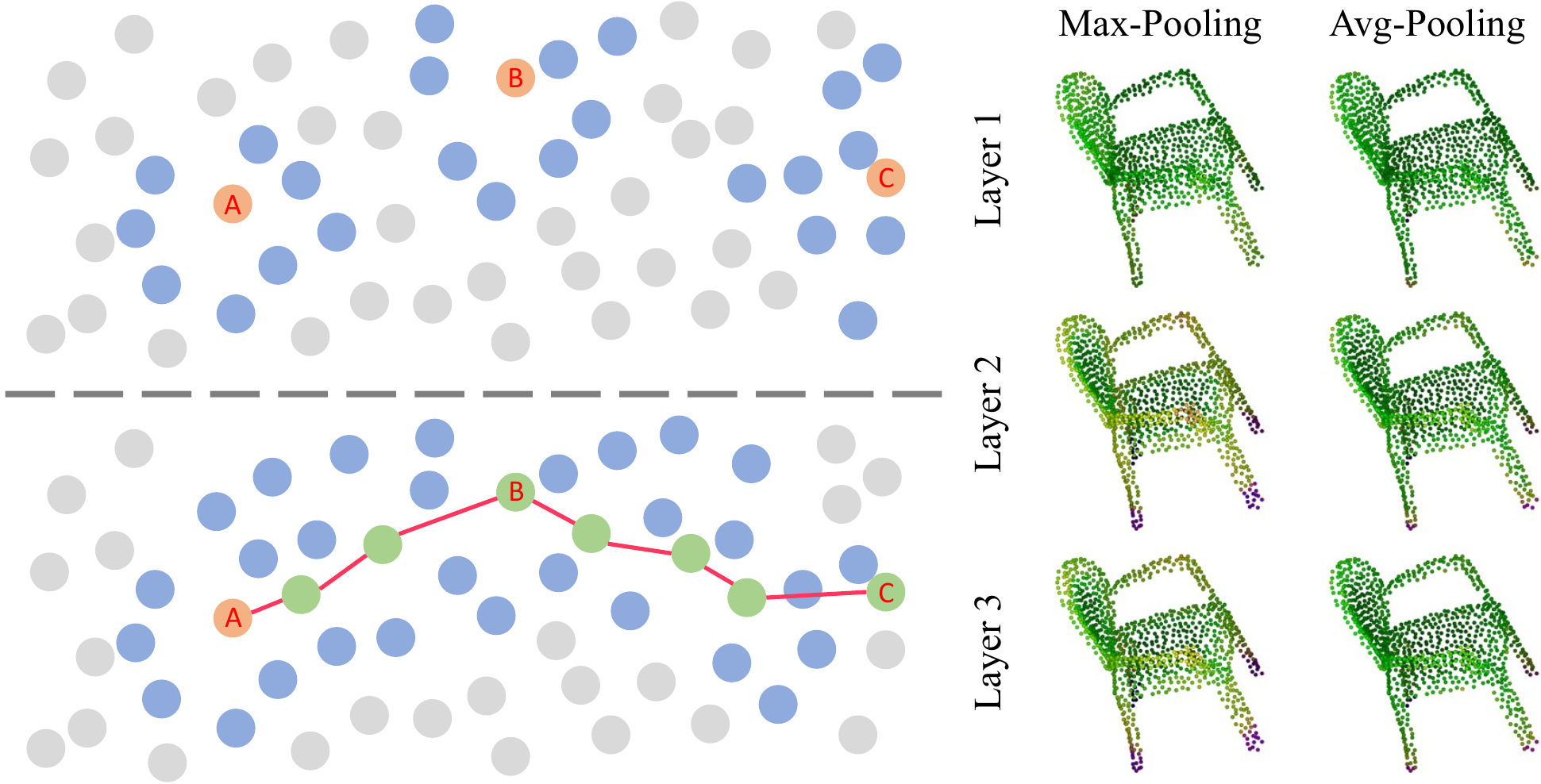}
	\end{center}
	\caption{\textbf{Left:} A point cloud projected on the 2D plane. \textbf{Left top:} Three key points and their k=9 query neighbors. \textbf{Left bottom:} A possible \textit{curve} through the three key points connected by red lines. \textit{Curve aggregation} fuses the features along the \textit{curve}. \textbf{Right:} Visualizations of the averaged channel $\mathbf{g}$ of a ModelNet40 chair object under different layers of local feature aggregation and different pooling strategies.}
	\label{fig:2}
\end{figure}

Using Manhattan distance $\Phi(\mathbf{f}, \mathbf{f}^j)=\mathbf{f} - \mathbf{f}^j$ as the relative encoding is the most natural practice and has been widely adopted. However, we argue that such encoding method does not provide abundant relative signals since most $\mathbf{g}$ in a point cloud contains nearly identical information in the same feature channel (regardless of pooling strategy), especially in shallow layers, as shown in Figure \ref{fig:2}. As the raw point cloud sampled from an explicit representation $\mathcal{R}$ of 3D objects is unordered, the point cloud $\mathbf{P}$ can be regarded as a set of random variables sampled in a particular probability density function (PDF) $\mathbf{U}$ modeled on $\mathcal{R}$, such that $\mathbf{P}\sim\mathbf{U}(\mathcal{R})$. 
After propagating through a certain number of network layers, $\mathbf{F}$ becomes the transformation over the random variable set $\mathbf{P}$.

We first consider an extreme case where $\mathbf{F}$ represents the initial point features, such that $\mathbf{F}=\mathbf{P}$. Using a simple 2D plane (Figure \ref{fig:2} left) as an exemplary $\mathcal{R}$, the sparse points are scattered on a compact subspace of $\mathbb{R}^2$. In Figure \ref{fig:2} left top, three key points are highlighted with their KNN computed neighbors. It can be practically observed that, after sampling the points, $\{\mathbf{p}_\text{A} - \mathbf{p}_\text{A}^j\}$ and $\{\mathbf{p}_\text{B} - \mathbf{p}_\text{B}^j\}$ are most likely to have similar values, as the key points are surrounded by their query neighbors in a similar pattern. However, point C at the boundary (an edge or an irregular segment of surface in 3D space) stands as an exception. Restricted by the geometry of $\mathcal{R}$, the distribution of point C query neighbors is considerably different than A and B, which leads to varying $\{\mathbf{p}_\text{C} - \mathbf{p}_\text{C}^j\}$. 

Based on the above observation, we claim that in any structured PDF that ensures the same sampling behaviour on similar geometries, $\mathbf{g}$ in Eq. \ref{la} is dependent on the distribution of $\mathbf{F}$ and $\mathbf{P}$. Point cloud objects, which have similar geometry information at most parts, turn out encoding similar and indistinguishable information in $\mathbf{g}$. As shown in Figure \ref{fig:2} right, the chair's back and seat have close features, especially in shallow layers. One possible strategy to enrich $\mathbf{g}$ is using more relative encoding rules rather than element-wise difference solely \cite{hu2020randla, bytyqi2020local}. In this paper, we enrich the local features $\mathbf{g}$ by combining the features aggregated from curves, as illustrated in Figure \ref{fig:2} left bottom. Each curve covers a long path in the point cloud encoding unique geometric information, which could be used to further increase point feature diversity.

\subsection{Curve Grouping} \label{cg}

In this subsection, we present the details on grouping curves in the feature space of point clouds. The starting point of a curve is essential to the overall grouping quality. To group $n$ curves simultaneously, the starting point set in $\mathbb{R}^{n\times ||\mathbf{f}||}$ needs to be determined beforehand. Borrowing the top-k selection method from \cite{gao2019graph}, we employ a sigmoid gated MLP to learn the selection score for each of the point features in $\mathbf{F}$. The starting points are the points with top $n$ scores. To enable gradient flow, we multiply the scores back to $\mathbf{F}$ via a self-attention manner.

After the construction of the starting point set, a walk $\mathcal{W}$ then starts from one of the starting points $\mathbf{s}_1$ and transits for exactly $l$ steps. The points traveled by $\mathcal{W}$ are grouped to form a curve $\mathbf{c}$. Given an intermediate state of curve $\mathbf{s}_i$ ($\mathbf{s}_i$ numerically equals $\mathbf{f}_i$ when grouping curves in feature space) arrived after walking for $i$ steps, we are interested in finding a walk policy $\pi(\mathbf{s}_i)$ that determines the next state of curve at the $i+1$ step. With a predefined $\pi(\cdot)$, a curve $\mathbf{c}=\{\mathbf{s}_1,\cdots,\mathbf{s}_l\}$ can be finally grouped by executing the following equation iteratively for $l$ times:
\begin{equation}
\begin{split}
    \mathbf{s}_{i+1} = \pi(\mathbf{s}_i), 1<=i\in \mathbb{Z}^+<=l.
\end{split}
\end{equation}


A good $\pi(\cdot)$ is essential to guarantee an effective curve grouping. Instead of a deterministic policy, we present a learnable $\pi(\cdot)$ that can be optimized along with the backbone network. In more detail, for a state $\mathbf{s}$, we apply MLPs on a \textit{state descriptor} $\mathbf{h}_{\mathbf{s}}\in \mathbb{R}^{2||\mathbf{s}||}$ to decide the next step. A state descriptor is constructed as the concatenation of point feature $\mathbf{s}_i$ and a \textit{curve descriptor} $\mathbf{r}_i$, which will be introduced later in this subsection. Selection logits $\alpha$ on all neighboring points in $N_\mathbf{s}$ can therefore be learned via the MLPs. We then feed $\alpha$ to a scoring function (e.g. softmax) to distribute each of the neighbors a score-based multiplier within $[0, 1]$. The point with the greatest score is then the output of our $\pi(\cdot)$. Formally, we formulate $\pi(\mathbf{s})$ as:

\begin{equation} \label{policy}
    \alpha = \{\texttt{MLP}(\mathbf{h}_{\mathbf{s}^j}|\mathbf{s}^j\in N_\mathbf{s})\},
\end{equation}
\begin{equation} \label{oldpolicy}
    \pi(\mathbf{s}) = \mathbf{F}[\argmax(\texttt{softmax}(\alpha))],
\end{equation}

\begin{figure}[t]
	\begin{center}
		\includegraphics[width=1\linewidth]{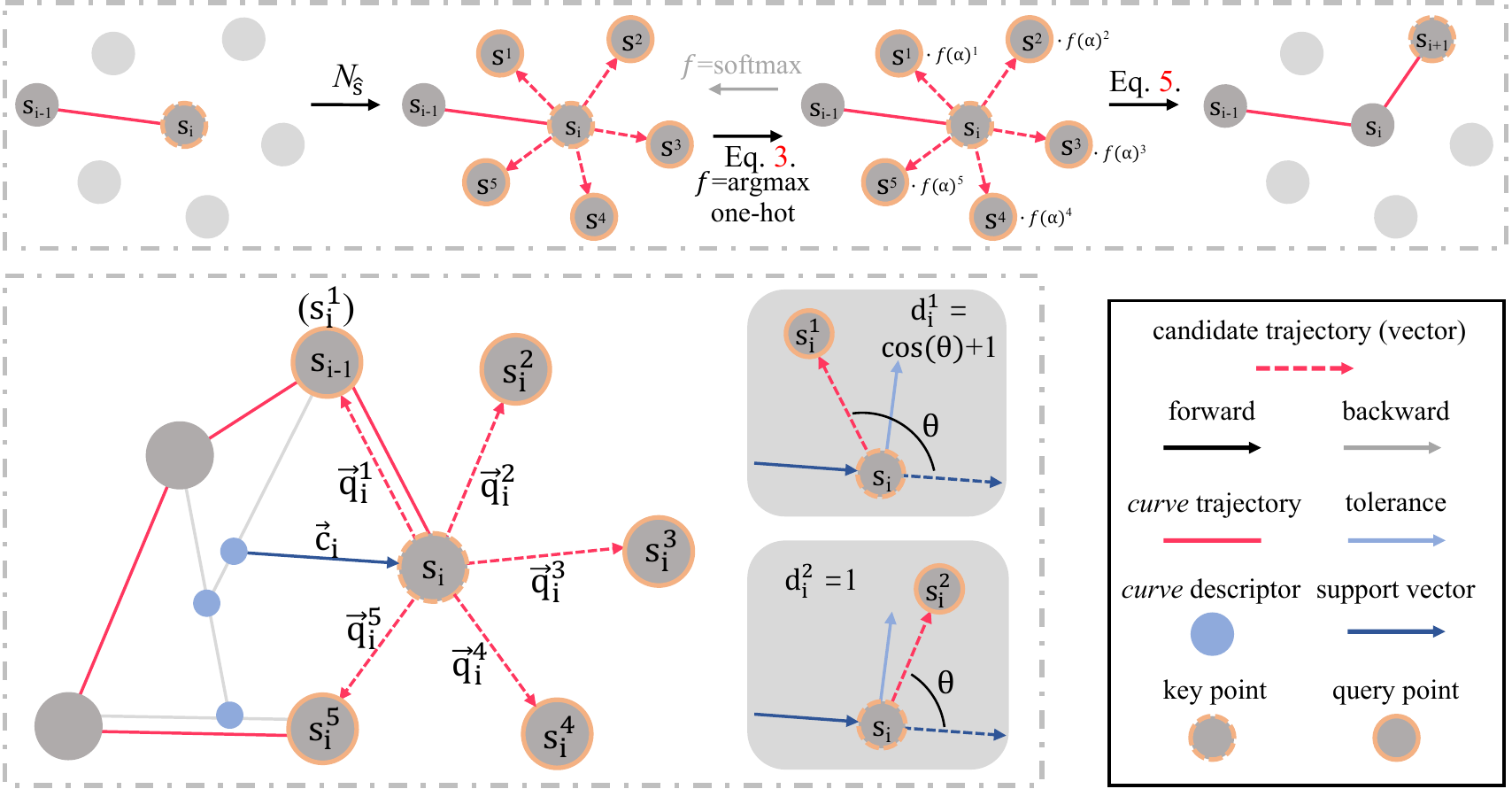}
	\end{center}
	\caption{\textbf{Top:} An overview of our curve grouping process. \textbf{Bottom:} Visualization of the proposed dynamic momentum and the crossover suppression strategies.}
	\label{fig:4}
\end{figure}


\noindent
where $\mathbf{h}_{\mathbf{s}^j}$ is the state descriptor of a KNN neighbor $\mathbf{s}^j$. In forward propagation, Eq. \ref{oldpolicy} determines the next state with the computed $\alpha$. However, during backward propagation, $\argmax$ obscures gradients, and hence the \texttt{MLP} in Eq. \ref{policy} is not able to be updated as expected.

Given the computed $\alpha$, we present an alternative equation to Eq. \ref{oldpolicy} that discards the $\argmax$ gate and enables gradient flow. First, we generate a hard one-hot style score vector for $\alpha$ instead of soft scores obtained from softmax function. By using gumbel-softmax \cite{jang2016categorical, maddison2016concrete, yang2019modeling} \footnote{Gumbel samplings are disabled for avoiding any randomness.} as the scoring function, logits can be converted into an one-hot vector based on the $\argmax$ index. Gradients through gumbel-softmax are computed identically to the ones through softmax. Then, we broadcastly multiply the query neighbors by the one-hot score vector and sum the multiplications together. The final result of the above operations is numerically identical to the ones computed by Eq. \ref{oldpolicy}. Consequently, our learnable policy is defined as follows:
\begin{equation}
    \pi(\mathbf{s}) = \sum_1^{k} ( \texttt{gumbel-softmax}(\alpha) \cdot {N_\mathbf{s}})),
\end{equation}
where $\cdot$ denotes broadcast multiplication along the feature dimension. An overview of the above pipeline is shown in Figure \ref{fig:4} bottom.

\begin{figure}[t]
	\begin{center}
		\includegraphics[width=1\linewidth]{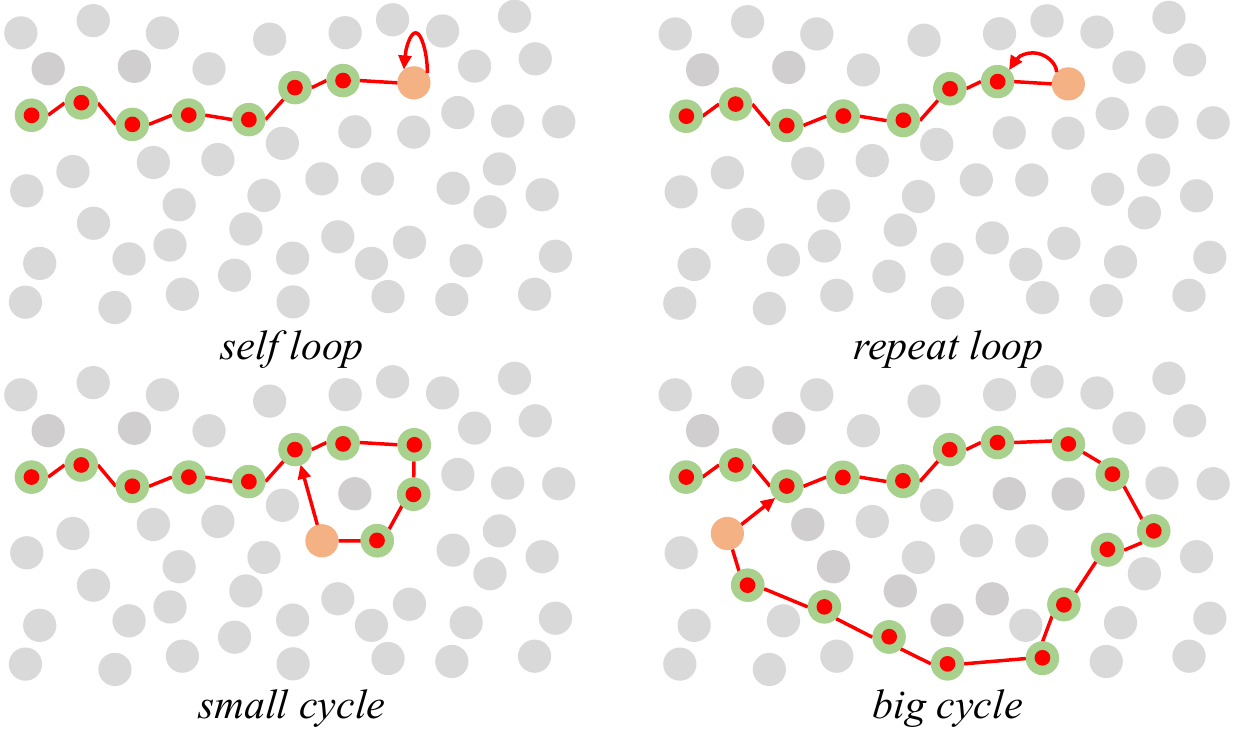}
	\end{center}
	\caption{\textbf{Four possible \textit{loops} in a curve}. The orange circle denotes current curve head, and the red arrow denotes current curve traveling direction.}
	\label{fig:3}
\end{figure}

By extending the curve with the highest score point, $\pi(\cdot)$ essentially determines the traveling direction of the curve based on the state descriptors in the neighborhood. We firstly follow a simple approach that constructs the state descriptor $\mathbf{h}_{\mathbf{s}^j}$ as the direct concatenation of $\mathbf{s}$ and the neighbor $\mathbf{s}^j$. However, such naive approach can easily lead to \textit{loops}, as Eq. \ref{policy} will always have the same output for the same input at each point. A \textit{loop} is a $\mathbf{c}=\{\mathbf{s}_1,\cdots,\mathbf{s}_l\}$ with repeated $\mathbf{s}$, which carries redundant and limited information, and hence should be avoided. Figure \ref{fig:3} shows four kinds of possible loops, among which \textit{self loop} can be easily avoided by excluding the key point itself during KNN computation. To avoid the other loops, the simple state descriptor formation will not suffice.

\noindent
\textbf{Dynamic momentum.} The key to avoid loops lies in a dynamic encoding of state descriptor with consideration of current curve progress. We maintain a \textit{curve descriptor} $\mathbf{r}_i\in \mathbb{R}^{||\mathbf{s}||}$ that encodes the prefix of the curve at step $i$. The state descriptor $\mathbf{h}_{\mathbf{s}^j}$ for each neighbor of the key point $\mathbf{s}_i$ now becomes the concatenation of $\mathbf{s}_i^j$ and $\mathbf{r}_i$.

Inspired by \cite{ioffe2015batch}, we update $\mathbf{r}_i$ following the momentum paradigm. However, we find that setting a fixed momentum coefficient $\beta$ is limited in terms of the final result. We propose that the prefix $\mathbf{r}$ of a curve can be better encoded by a dynamic momentum variant, such that:
\begin{equation}
\begin{split}
    \beta &= \texttt{softmax}(\texttt{MLP}([\mathbf{r}_{i-1}, \mathbf{s}_i])),\\
    \mathbf{r}_i &= \beta \mathbf{r}_{i-1} + (1-\beta) \mathbf{s}_i,
\end{split}
\end{equation}
where $[\cdot]$ represents concatenation. Figure \ref{fig:4} bottom illustrates the dynamic momentum paradigm.


\begin{figure*}[t]
	\begin{center}
		\includegraphics[width=0.9\linewidth]{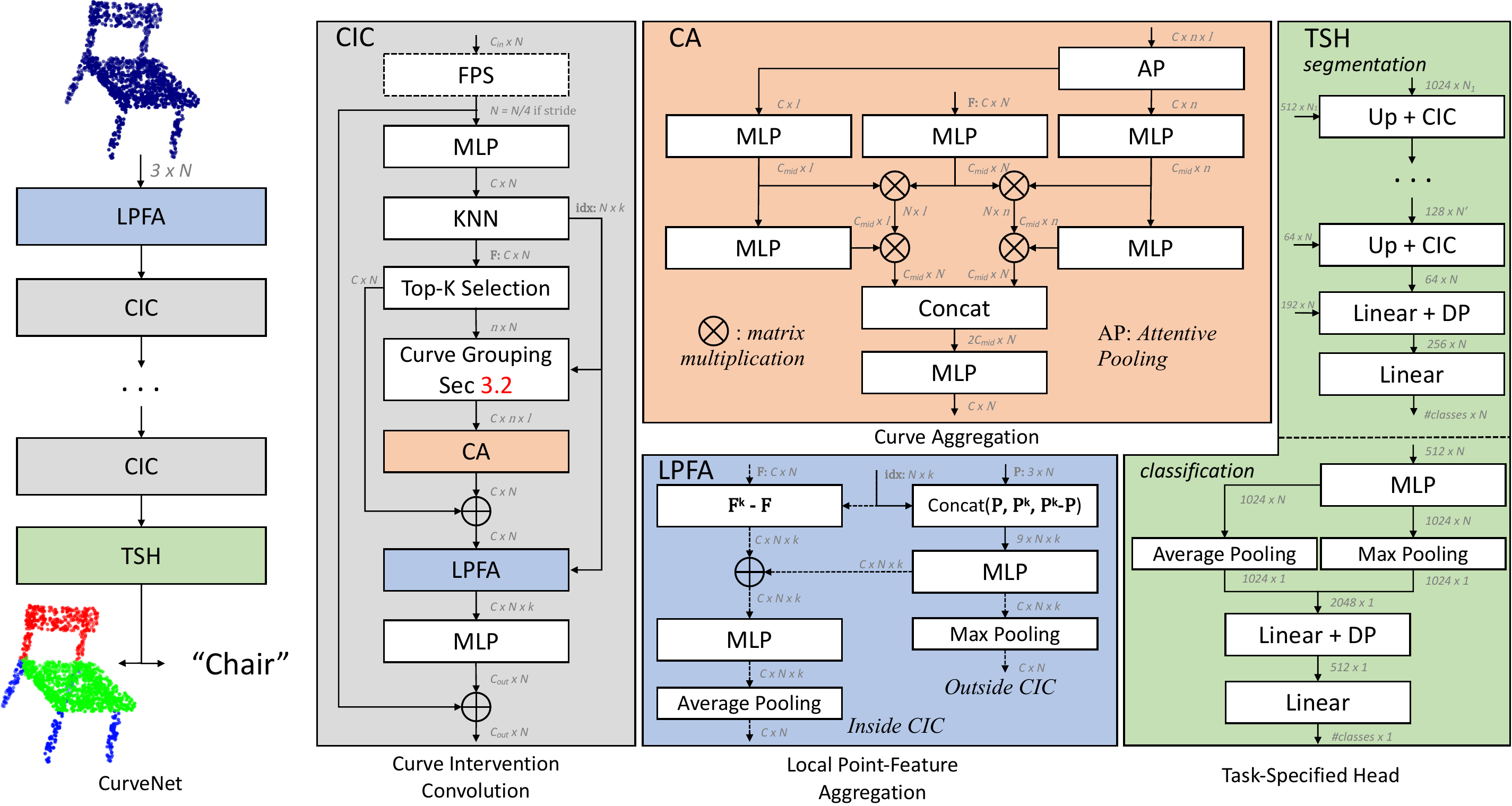}
	\end{center}
	\caption{\textbf{CurveNet overview}. The network is comprised of a stack of building blocks. FPS denotes the farthest point sampling method \cite{qi2017pointnet++}. Dotted blocks and lines are optional regarding different blocks. Building blocks are matched in abbreviation and color.}
	\label{fig:network}
\end{figure*}

\noindent
\textbf{Crossover suppression.} Although the dynamic momentum strategy avoids loops, curves may still encounter \textit{crossovers}. Unlike loops, small number of crossovers may imply useful patterns and should not be avoided completely. However, when a large number of crossovers occurs, same node will be included repeatedly, hurting the curve representation. Therefore, we propose to suppress crossovers by investigating the curve's traveling direction. 

We first construct a \textit{support vector} $\overrightarrow{\mathbf{c}}_i=\mathbf{s}_i-\mathbf{r}_{i-1}$ representing the general direction of the current curve at step $i$. Subsequently, for each query neighbor of the curve head $\mathbf{s}_i$, we compute the \textit{candidate vector} as $\overrightarrow{\mathbf{q}}_i^j=\mathbf{s}_i^j-\mathbf{s}_i$. The included angle $\theta$ between $\overrightarrow{\mathbf{c}}$ and $\overrightarrow{\mathbf{q}}^j$ indicates whether the curve is about to go straight or make a turn. We suppress crossovers by scaling down $\alpha^j$ that has large $\theta$ (i.e. likely to turn around and cause potential crossovers).

Specifically, we determine the angle distance between $\overrightarrow{\mathbf{c}}$ and $\overrightarrow{\mathbf{q}}^j$ through cosine similarity and the measurement for each vector pair strictly falls into the range $[-1, 1]$. A value close to $-1$ represents the two vectors are in the opposite direction (should be suppressed) and 1 represents they are in the same direction (should not be suppressed). Considering there exist boundaries in the latent space, a curve cannot go straight forever without making a turn. We therefore set a tolerance threshold angle $\bar{\theta}$ \footnote{Learning such value yields poorer results in our experiments.}, so that only the candidate vector with included angle greater than $\bar{\theta}$ needs to be suppressed. Following the above intuition, we then construct the \textit{crossover suppression multiplier} $\mathbf{d}^j$ by shifting and clipping the cosine similarity score into [0, 1]. Potential crossovers are suppressed by scaling the candidate logit $\alpha^j$ with $\mathbf{d}^j$. An illustration of the crossover suppression strategy is outlined in Figure \ref{fig:4} bottom.


\subsection{Curve Aggregation and CurveNet} \label{ca}

As discussed in Sec \ref{rethink}, the purpose of curve aggregation is to enrich the intra-channel feature variety of the relative encodings $\phi$ and eventually provides a better description to $\mathbf{g}$. For notation simplicity, we define the number of feature channels as $C$, the number of points as $P$, and a basic Attentive-Pooling operator \cite{hu2020randla} as AP. In AP, the input features $\in \mathbb{R}^{C\times\ast}$ are scaled in the self-attention style and summed up along the $\ast$ dimension, resulting in $\mathbb{R}^{C\times 1}$.


Given the grouped curves $\mathbf{C}=\{\mathbf{c}_1,\cdots,\mathbf{c}_n\}\in \mathbb{R}^{C\times n\times l}$, to aggregate the features in curves, we consider both inter-relations among the curves and intra-relations within each curve. We first learn an inter-curve feature vector $f_{inter}\in \mathbb{R}^{C\times l}$ and an intra-curve feature vector $f_{intra}\in \mathbb{R}^{C\times n}$ by applying AP on $\mathbf{C}$ along different axis. Point features $\mathbf{F}$ together with $f_{intra}$ and $f_{inter}$ are then fed to three individual bottleneck MLPs to reduce feature dimensions. We apply matrix multiplication on the reduced $\mathbf{F}$ with reduced $f_{intra}$ and $f_{inter}$ separately to learn the respective curve-point mappings. Softmax functions are employed to convert the mappings into scores. In another branch, the reduced $f_{intra}$ and $f_{inter}$ are further transformed with two extra MLPs, which are then fused with the computed mapping scores by matrix multiplication separately. The above process ends up with two fine-grained feature vectors $f'_{intra}$ and $f'_{inter}$ in the same shape of $\mathbb{R}^{C\times P}$. We concatenate $f'_{intra}$ and $f'_{inter}$ along the feature axis and feed into a final MLP. The output of our curve aggregation is the residual addition to the origin input.

We embed our Curve Grouping (CG) block and Curve Aggregation (CA) block into a Curve Intervention Convolution (CIC) block. In each CIC block, curves are first grouped (Sec \ref{cg}) and then aggregated to all point features (Sec \ref{ca}). We stack 8 CIC blocks together to construct a ResNet \cite{he2016deep} style network, referred to as CurveNet. Our CurveNet initially learns a relative local encoding of the input point coordinates through the Local Point-Feature Aggregation (LPFA) block, which projects the relative point difference into a higher dimension. CurveNet eventually makes predictions through the Task-Specified Head (TSH) regarding of different point cloud processing tasks. For classification task, the extracted point features are first pooled and then passed into two fully-connected layers. For segmentation task, we use an attention U-Net \cite{oktay2018attention} style decoder that concatenates attentive shortcut connections from the encoder. Figure \ref{fig:network} gives an overview of our CurveNet. Network structure and building block details are presented in the supplemental materials.





		

\section{Experiments} \label{exp}
We present experimental results for our point cloud object analysis methods on object classification, shape part segmentation, and normal estimation tasks. 


\subsection{Implementation Details}
In all experiments, Dropout layers \cite{srivastava2014dropout} were adopted in final linear layers with probability 0.5 \cite{wang2019dynamic}. We used LeakyReLU as the activation function in the backbone sub-network and ReLU in the task-specific heads. $\bar{\theta}$ was set to $90^\circ$. To eliminate the influence of randomness, random seed was fixed in all experiments, which were implemented in the PyTorch framework \cite{paszke2019pytorch}.

For classification tasks, we used SGD with momentum 0.9 as the optimizer and set the number of neighbors in KNN to 20. For segmentation tasks, the number of KNN neighbors was set dynamically according to different radius with no more than 32. The point features were interpolated similar to \cite{qi2017pointnet} during upsampling. The distances between predictions and ground truth labels were minimized through cross entropy loss.

\subsection{Benchmarks}

\begin{table} 
	\begin{center}
	\caption{ModelNet10 (M10) and ModelNet40
(M40) classification accuracy (\%). `nr' denotes using normal vectors as extra inputs. `*' denotes methods evaluated with voting strategy \cite{liu2019relation}. }
		\begin{tabular}{l|cccc} 
			\toprule 
			Methods & Input & \#point & M10$\uparrow$ & M40$\uparrow$\\
			\hline
			\hline
			PointNet \cite{qi2017pointnet} & xyz & 1024 & - & 89.2\\
			PointNet++ \cite{qi2017pointnet++} & xyz & 1024 & - & 90.7\\
			DGCNN \cite{wang2019dynamic} & xyz & 1024 & - & 92.9 \\
			PointASNL \cite{yan2020pointasnl} & xyz & 1024 & 95.7 & 92.9\\
			RS-CNN \cite{liu2019relation} & xyz & 1024 & - & 92.9 \\
			RS-CNN \cite{liu2019relation} * & xyz & 1024 & - & 93.6 \\
			Grid-CNN \cite{xu2020grid} & xyz & 1024 &  \textbf{97.5} & 93.1 \\
			PCT \cite{guo2020pct} & xyz & 1024 & - & 93.2 \\
			PAConv \cite{xu2021paconv} & xyz & 1024 & - & 93.6 \\
			PAConv \cite{xu2021paconv} * & xyz & 1024 & - & 93.9 \\
			\hline
			A-CNN \cite{komarichev2019cnn} & xyz, nr & 1024 & 95.5 & 92.6\\
			PosPool \cite{liu2020closer} *& xyz & 2048 & - & 93.2 \\
			
			SO-Net \cite{li2018so} & xyz, nr & 5000 & 95.7 & 93.4\\
			\hline
			CurveNet (Ours) & xyz & 1024 & 96.1 & \textbf{93.8}\\
			CurveNet (Ours) * & xyz & 1024 & 96.3 & \textbf{94.2}\\
			\bottomrule
		\end{tabular}
		
		\label{table:modelnet}

	\end{center}
	\vspace{-1em}
\end{table} 

\begin{table} 
	\begin{center}
	\caption{ShapeNet part results in mean intersection of union (\%).}
		\begin{tabular}{l|ccc} 
			\toprule 
			Methods & Input & \#point & mIoU$\uparrow$\\
			\hline
			\hline
			PointNet \cite{qi2017pointnet} & xyz & 2048 & 83.7\\
			DGCNN \cite{wang2019dynamic} & xyz & 2048  & 85.1 \\
			PointCNN \cite{li2018pointcnn} * & xyz & 2048 & 86.1 \\ 
			PointASNL \cite{yan2020pointasnl} & xyz & 2048 & 86.1\\
			RS-CNN \cite{liu2019relation} *& xyz & 2048 & 86.2 \\
			PAConv \cite{xu2021paconv} & xyz & 2048 & 86.0 \\
			PAConv \cite{xu2021paconv} * & xyz & 2048 & 86.1 \\
			PCT \cite{guo2020pct} * & xyz & 2048 & 86.4 \\
			
			\hline
			PointNet++ \cite{qi2017pointnet++} & xyz, nr & 2048 & 85.1 \\
			SO-Net \cite{li2018so} & xyz, nr & 1024 & 84.6\\
			\hline
			CurveNet w/o curves (Ours) & xyz & 2048 & 85.9\\
			CurveNet (Ours)  & xyz & 2048 & \textbf{86.6}\\
			CurveNet (Ours) *& xyz & 2048  & \textbf{86.8}\\
			\bottomrule
		\end{tabular}
		
		\label{table:shapenet}

	\end{center}
	\vspace{-1em}
\end{table} 

\begin{table} 
	\begin{center}
	\caption{Normal estimation results in avg cosine-distance error.}
		\begin{tabular}{l|ccc} 
			\toprule 
			Methods & Input & \#point & Error$\downarrow$\\
			\hline
			\hline
			PointNet \cite{qi2017pointnet} & xyz & 1024 & 0.47\\
			DGCNN \cite{wang2019dynamic} & xyz & 1024  & 0.29 \\
			
			RS-CNN \cite{liu2019relation} & xyz & 1024 & 0.15 \\
			PCT \cite{guo2020pct}  & xyz & 1024  & 0.13 \\
			\hline
			CurveNet w/o curves (Ours) & xyz & 1024 & 0.16\\
			CurveNet (Ours) & xyz & 1024 &  \textbf{0.11}\\
			\bottomrule
		\end{tabular}
		
		\label{table:normal}

	\end{center}
	\vspace{-2em}
\end{table} 

		


\noindent
\textbf{Object classification.} ModelNet 10/40 datasets \cite{wu20153d} are the most commonly used datasets for object shape classification benchmarks, which collect meshed CAD models across a variety of objects. ModelNet10 dataset is comprised of 4899 individual models that are distributed into 10 different categories. We split the training and testing samples following the same schema as in \cite{liu2019point2sequence}. In a larger homogeneous dataset, ModelNet40 consists of 12311 models that are classified into 40 categories. In both datasets, we only used the coordinates of 1024 uniformly sampled points as network inputs. The points were normalized into unit spheres before feeding into networks. A random scaling multiplier within the range [0.66, 1.5] was first multiplied on the sampled points. Then, each point was translated along the three directions by random displacements within [-0.2, 0.2]. The scaling and translation settings were consistent to the ones used in \cite{klokov2017escape, liu2019relation}. We trained the models for 200 epochs, starting with a learning rate of 0.1 and cosineannealling scheduled to 0.001 in 200 epochs. Batch size was set to 32 for training, and 16 for validation.

\begin{figure}[t]
	\begin{center}
		\includegraphics[width=0.9\linewidth]{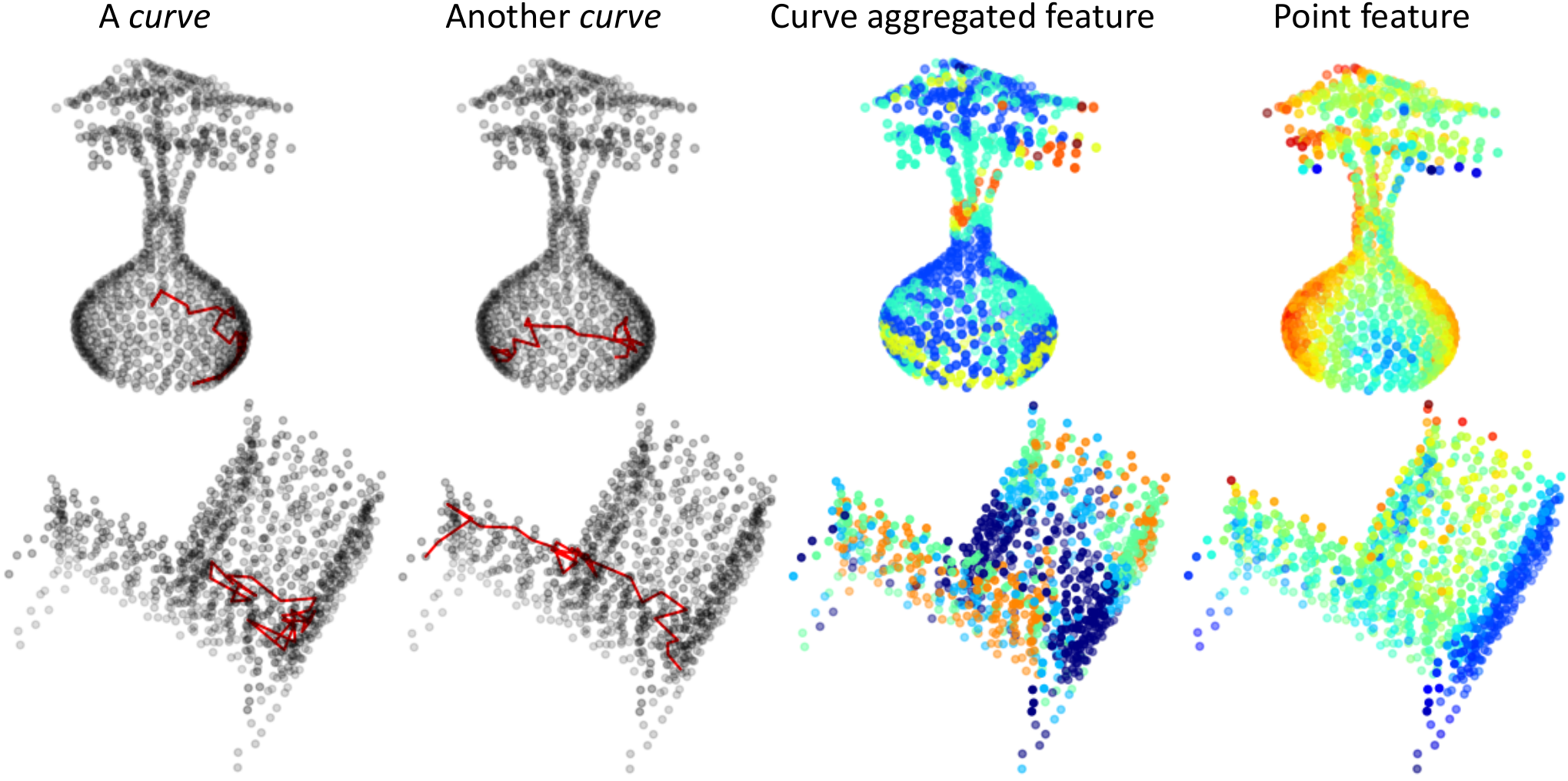}
	\end{center}
	\caption{\textbf{Visualizations of curves and curve features.} \textbf{Left:} Loops are avoided and crossovers are suppressed in the grouped curves. \textbf{Right:} Point features can be enriched by combining the long-range curve features.}
	\label{fig:curves}
\end{figure}

Table \ref{table:modelnet} reports the comparison results between our CurveNet and the most recent methods. With only 1024 uniformly sampled points, our method achieved the state-of-the-art result on the large scale ModelNet40 dataset at 93.8\% without voting \cite{liu2019relation} and reached 94.2\% when averages 10 prediction votes. We also achieved 96.3\% on the ModelNet10 subset, which is the second best result among all methods with the same training data. Moreover, CurveNet is a highly memory-efficient architecture that requires only 4.1 G GPU memory for training while DGCNN \cite{wang2019dynamic} requires 5.9 G. We visualize the grouped curves, the local aggregated features, and the curve aggregated features on two randomly selected channels in Figure \ref{fig:curves}. The curves are able to cover long-range semantics and hence bring channel diversity to a great extent.



\noindent
\textbf{Object part segmentation.} We validated our method on the ShapeNetPart dataset \cite{yi2016scalable} for the 3D shape part segmentation task. The dataset collects 16881 shape models across 16 categories. Most objects in the dataset have been labeled with less than 6 parts, results in a total number of 50 different parts. Our training and testing split scheme follows \cite{qi2017pointnet, qi2017pointnet++}, such that 12137 models were used as training samples while the rest were used as validation. 2048 points were uniformly sampled from each model to be the input to our networks. We trained the models for 200 epochs with batch size 32, starting with a learning rate of 0.05 which decayed by 0.1 at the 140th and 180th epoch. The momentum and weight decay are set to 0.9 and 0.0001, respectively. We inserted a simple SE \cite{hu2018squeeze} module before the last linear layer of our CurveNet. Same to \cite{liu2019relation, bytyqi2020local}, the one-hot class label vector and global feature vectors are also adopted.

\begin{figure}[t]
	\begin{center}
		\includegraphics[width=1.0\linewidth]{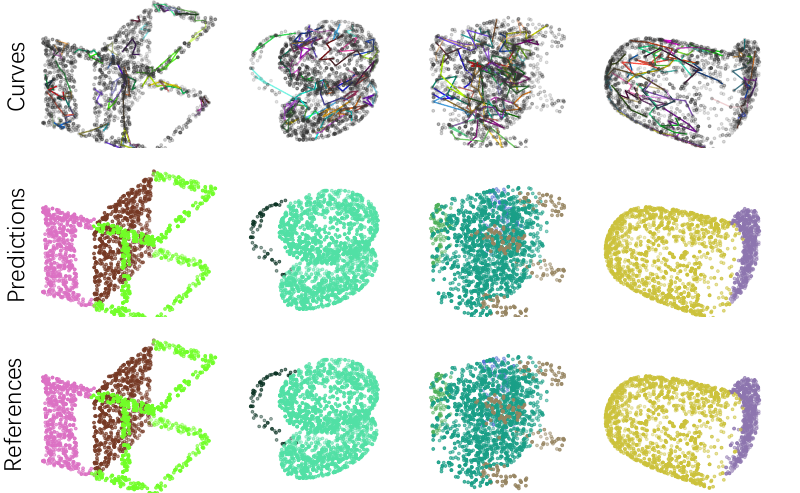}
	\end{center}
	\caption{\textbf{Visualizations of curves and segmentation results.} Random selected curves are plotted with random colors.}
	\label{fig:shapenet}
\end{figure}

Mean intersection of union (mIoU) results across instances are reported in Table \ref{table:shapenet}, and the category-wise mIoU scores are presented in the supplemental materials. Our method achieves state-of-the-art overall mIoU of 86.6\%, surpasses all existing methods. Without grouping any curves, our base architecture reaches 85.9\%, proving the effectiveness of involving curves in point cloud shape segmentation task. Moreover, we visualized four cases qualitatively along with the learned curves in Figure \ref{fig:shapenet}. The grouped curves were able to explore both short and long range shape relations. Model complexity is reported and analyzed in the supplementary materials.

\noindent
\textbf{Object normal estimation.} Object surface normal is essential to 3D modeling and rendering. Differing from understanding objects part by part, estimating normal requires a comprehensive understanding of the entire object shape and geometry. We validate our CurveNet on estimating normal by using ModelNet40 dataset, where each point in the point cloud has been labeled with its three-directional normal. CurveNet architecture is constructed similarly to the one used in segmentation task, excluding the one-hot class label vector and global feature vectors. Models are trained with an initial learning rate of 0.05 and cosineanealing scheduled to 0.0005 in 200 epochs.

Table \ref{table:normal} shows the average cosine-distance error comparisons of CurveNet and state-of-the-art methods. Without any curves, our base CurveNet architecture achieves an average error of 0.16, which is closed to \cite{liu2019relation, guo2020pct}. When curves are involved, our full CurveNet demonstrates a superior performance with 0.11 average error, setting a new benchmark to the normal estimation task.


\subsection{Ablation Studies}

We conducted extensive experiments on ModelNet40 dataset to study our proposed method comprehensively. Unless explicitly specified, implementation details remained the same as the ones described in the benchmark section. All the ablative studies were examined without voting.

\noindent
\textbf{Component studies.} The impact of individual component of CurveNet was examined by simply removing or replacing them from the full CurveNet architecture. We conducted experiments on replacing LPFA with the common local feature aggregation as in Eq. \ref{la}, disabling the dynamic momentum and the crossover suppression strategies, and replacing the proposed CA operator to the vanilla non-local module (i.e. intra-relations and inter-relations are not separated). The results are reported in Table \ref{comp}. 

We observed that, although using LPFA solely cannot bring significant performance improvement (models A and B), with the intervention of curves, the presence of LPFA is able to make a huge difference in terms of the classification result (models F and G). As shown in models C and D, both the proposed dynamic momentum and crossover suppression strategy are empirically effective as expected. Moreover, from model E, we find that the proposed curve aggregation operator plays the most significant role in the CurveNet. When aggregating grouped curve features through the vanilla non-local module, the accuracy is dropped by 0.7\%, with no benefit on the inference latency.  


\begin{table}[t]
	\begin{center}
	\caption{Component study results. Top-1 classification accuracy and per point cloud (per batch) inference time are reported. LPFA: the Local Point-Feature Aggregation, CG: the Curve Grouping operator, DM: the Dynamic Momentum strategy, CS: the Crossover Suppression strategy, and CA: the Curve Aggregation operator.}
	    
		\resizebox{\columnwidth}{!}{\begin{tabular}{c|ccc c|c c} 
			\toprule 
			& & \multicolumn{2}{c}{CG} &&\\\cline{3-4}
			Model & LPFA & DM & CS & CA & Acc (\%) & latency (ms) \\
			\hline
			\hline
			A& & & & &93.1&37.6(140)\\
			B& \checkmark & & & &93.3 &37.5(143)\\
			C& \checkmark & \checkmark & & \checkmark& 93.4&44.3(145)\\
			D & \checkmark &  & \checkmark& \checkmark& 93.3&44.1(144)\\
			E & \checkmark & \checkmark& \checkmark& & 93.1&45.0(146)\\
			F &  & \checkmark & \checkmark&\checkmark &93.4 &44.8(143)\\
			G & \checkmark & \checkmark& \checkmark&\checkmark &93.8  &45.2(146)\\
			\bottomrule
		\end{tabular}}
		
		\label{comp}
	\end{center}
\end{table} 
\begin{figure}[t]
	\begin{center}
		\includegraphics[width=1\linewidth]{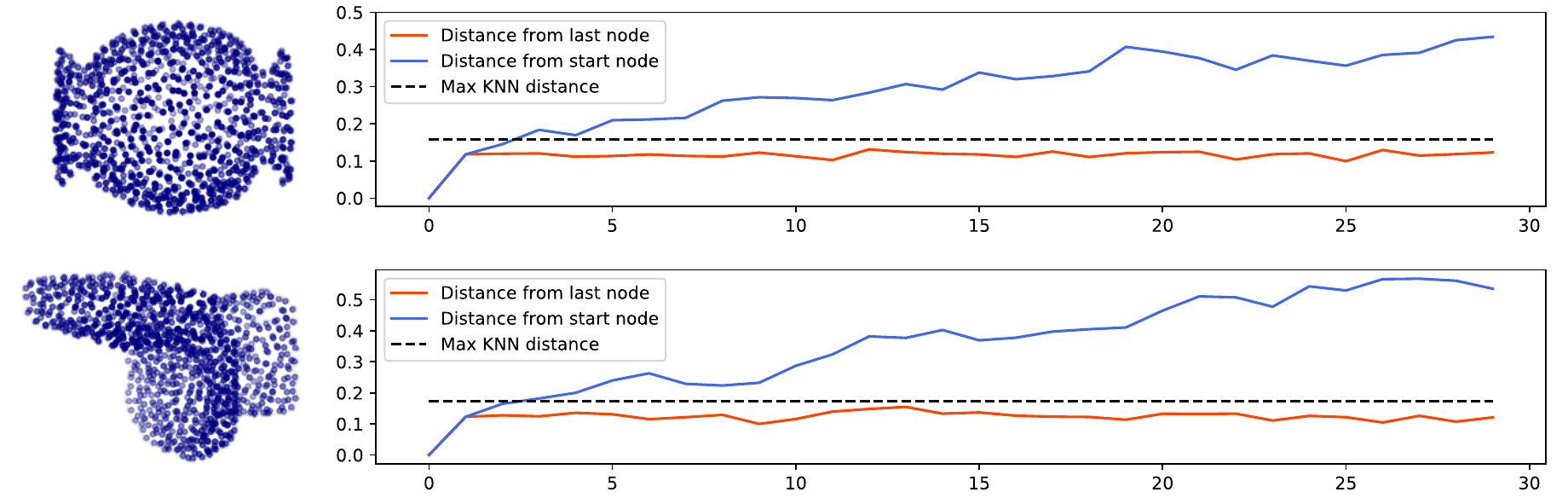}
	\end{center}
	\caption{\textbf{How far do the curves go?} Left are two objects with distinct shapes. Right are the average euclidean distances to the start node and the last node of all curves with 30 steps in group 1.}
	\label{fig:travel}
\end{figure}


\noindent
\textbf{Shallow layer vs deep layer.} In Sec. \ref{rethink}, we claimed that shallow features after local aggregation lack single channel diversity, and curve features could be more desired at shallow layers than at deep layers of a network. We conducted experiments on aggregating curves with various quantities and lengths on different groups of our CurveNet, the results are shown in the supplemental materials (Figure 1). Aggregating curves at shallow layers (group 1/2) yield better results than at deep layers (group 3/4), which proved our claim empirically. 

	    
		

\noindent
\textbf{Curve quantity vs curve length.} Curve quantity $n$ and length $l$ are the two hyper parameters determining the network performance directly. Short curves cannot capture long-range patterns, while long curves require better guidance and may contain redundant information. To study the relations between curve quantity and curve length, we conducted experiments on fixed total point number in curves. Figure 1 bottom right in the supplemental materials shows that although a long curve (length 50) is able to achieve the best result, the network performance degrades as the curve extends further. Aggregating longer curves is also computational inefficient, as the transition of nodes cannot be computed parallelly. To verify whether the curves are trapped in a local region, we present the average Euclidean distances between each node of the curves to the starting/last point in Figure \ref{fig:travel}. The curves are capable of jumping out of the maximum local KNN range to explore longer range relations.

\begin{figure}[t]
	\begin{center}
		\includegraphics[width=1\linewidth]{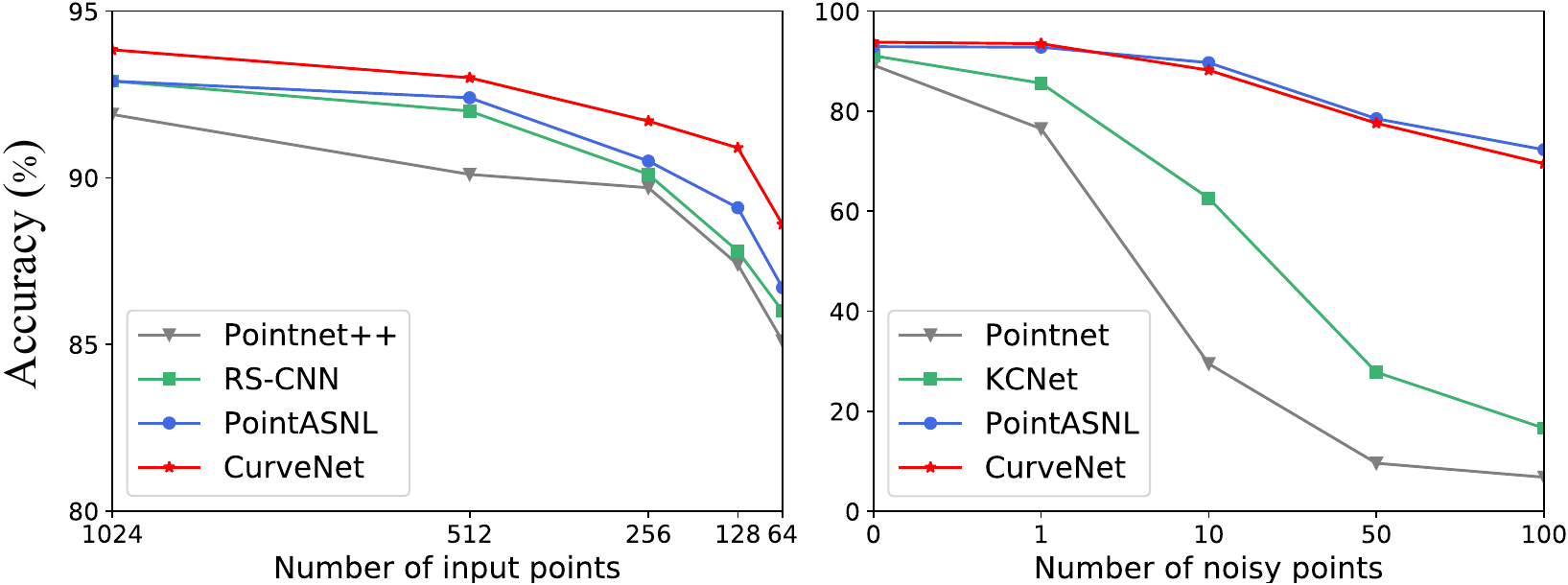}
	\end{center}
	\caption{\textbf{Left:} Comparison on sparser training and testing inputs. \textbf{Right:} Comparison on noisy inputs, with voting enabled.}
	\label{fig:robust}
\end{figure}

\noindent
\textbf{Sparser input points and noisy testing points.} Curve grouping could be sensitive to point cloud sparsity and noise. We conduct extensive experiments on \textbf{(1)} training and testing with sparser input points and \textbf{(2)} training on 1024-point raw coordinates and testing with noisy points \cite{yan2020pointasnl}. As shown in Figure \ref{fig:robust} left, our CurveNet achieves the best results on all experiments regarding of different number of input points. For noise tests, we add an extra max pooling layer following the first LPFA block. Our CurveNet outperforms \cite{shen2018mining, qi2017pointnet} in all experiments and achieves on par results to \cite{yan2020pointasnl}, demonstrating the robustness to noise.


\section{Conclusion}
In this paper, we proposed a long-range feature aggregation method, namely \textit{curve aggregation}, for point clouds shape analysis. We first discussed the potential drawbacks of the existing local feature aggregation paradigm, and claimed the need for the aggregation of point cloud geometry. We then presented our method in two sequential steps: the rules for grouping curves in a point cloud and the integration of the grouped curve features with the extracted point features. During the process, potential problems were defined and resolved. Our method achieved state-of-the-art results on multiple point clouds object analysis tasks.

{\small
\bibliographystyle{ieee_fullname}
\bibliography{egbib}
}

\end{document}